\documentclass[11pt,a4paper]{article}
\usepackage[T1]{fontenc}
\usepackage[utf8]{inputenc}
\usepackage{lmodern}
\usepackage{amsmath,amssymb}
\usepackage[margin=24mm]{geometry}
\IfFileExists{microtype.sty}{\usepackage{microtype}}{}
\usepackage{graphicx}
\usepackage{booktabs,longtable,array,calc}
\usepackage{etoolbox}
\usepackage{float,needspace,placeins}
\usepackage[font=small,labelfont=bf,labelsep=quad]{caption}
\usepackage{xurl}
\usepackage[numbers,sort&compress]{natbib}
\usepackage[unicode,colorlinks=true,linkcolor=black,citecolor=black,urlcolor=blue]{hyperref}
\usepackage{bookmark}

\AtBeginEnvironment{longtable}{\small\setlength{\tabcolsep}{4pt}}

\makeatletter
\def\maxwidth{\ifdim\Gin@nat@width>\linewidth\linewidth\else\Gin@nat@width\fi}
\def\maxheight{\ifdim\Gin@nat@height>0.82\textheight0.82\textheight\else\Gin@nat@height\fi}
\setkeys{Gin}{width=\maxwidth,height=\maxheight,keepaspectratio}
\def\fps@figure{H}
\makeatother
\hypersetup{pdftitle={SignTrace: Describe a Sign, Find the
Word},pdfauthor={Zengji Tu; Xingye Zhu; Ningjing Wang; Tingyi Huang; Yangjunfeng Zhu; Dai Wan}}
\begin{document}
\begin{center}
{\fontsize{18}{22}\selectfont\bfseries SignTrace: Describe a Sign, Find
the Word\par}
\vspace{10pt}
Zengji Tu\textsuperscript{1,2,*}, Xingye Zhu\textsuperscript{1,2}, Ningjing Wang\textsuperscript{1,2},\\[2pt]
Tingyi Huang\textsuperscript{1,2}, Yangjunfeng Zhu\textsuperscript{1,2}, Dai Wan\textsuperscript{3}\\[7pt]
{\small\textsuperscript{1}Peking University, Beijing 100871, China\\
\textsuperscript{2}Sign Language Department, Loving Heart Society, Peking University,\\
Beijing 100871, China\\
\textsuperscript{3}Institute of Population Research, Peking University, Beijing 100871, China\\[4pt]
\textsuperscript{*}Corresponding author: \href{mailto:tuzengji@stu.pku.edu.cn}{tuzengji@stu.pku.edu.cn}\\[6pt]
September 13, 2026}
\end{center}
\begin{abstract}
Identifying an unfamiliar sign is difficult when a learner remembers its
movement but does not know its meaning or formal feature codes.
SignTrace addresses this longstanding reverse-lookup problem through
natural-language access to a Chinese sign-language dictionary. The
system integrates LLM-based dictionary enrichment, action extraction,
dictionary-style rewriting, seven-channel retrieval, and candidate
reranking over 6,699 entries. It has been deployed for user trials and
has received positive informal feedback. Evaluation on a
dictionary-derived benchmark of 500 movement-description queries yields
94.0\% Hit@1, 97.4\% Hit@9, and a mean reciprocal rank of 0.9540.
Reranking increases Hit@1 from 71.8\% to 94.0\%, while component
analyses show the contribution of enriched entry descriptions. Median
query-processing time is 13.37 seconds with six concurrent queries. By
connecting everyday movement descriptions to documented signs and
meanings, SignTrace provides a practical tool for identifying unfamiliar
signs. Dictionary-derived wording and prior selection within the
benchmark limit generalization to descriptions independently produced by
users.
\end{abstract}
\vspace{6pt}
\section{1 Introduction}\label{introduction}

A sign language learner may remember the fingertips of one hand touching
the mouth, followed by a downward movement as the fingers spread, yet
not know the corresponding word. A conventional word-to-sign lookup
cannot begin with this observation alone. This creates a practical
barrier to identifying unfamiliar signs: the information the learner has
is an action, while the dictionary's usual entry point is a known word.

Sign language dictionaries link documented forms to written meanings and
instructional material. Feature-based dictionaries, recognition, and
hybrid search offer alternatives to word-based access
\citep{aslsearch2015, hybrid2022, reduced2025}. Free-text reverse lookup
provides another route: learners describe what they remember in ordinary
language, without first selecting formal feature codes or recording a
performance. The returned dictionary entries supply meanings and
illustrations that can be checked against the observation.

The technical difficulty is that observers and dictionaries often
express the same movement differently. A novice may write that a hand
makes an L shape or that two fingers are pinched. A dictionary may
specify finger extension, palm orientation, body location, contact, and
the temporal order of movements. A short query can omit the feature that
distinguishes two otherwise similar signs. A plausible guess of the
intended word may also contradict the described action. Effective
reverse lookup must connect these different descriptions while retaining
the details needed to distinguish candidates.

We introduce \textbf{SignTrace}, a system that coordinates LLMs across
dictionary preparation, query interpretation, and candidate comparison.
Enriched entry descriptions make action details searchable. Parallel
action extraction and dictionary-style rewriting connect informal
observations to that resource, while preserving the original query. Five
textual channels, one structural channel, and one meaning-hint channel
retrieve candidates; an LLM then scores their agreement with the
extracted observation. Together, these stages allow learners to search
the dictionary using natural-language movement descriptions.

SignTrace combines an operational lookup service with an empirical
evaluation of its retrieval architecture. The service has been released
for user trials and has received positive informal feedback. Evaluation
on 500 dictionary-derived queries yields 94.0\% Hit@1 and 97.4\% Hit@9
over 6,699 entries. Baseline comparisons, component analyses, and error
analysis characterize the contributions of dictionary enrichment and
reranking, together with the remaining retrieval limitations.

\section{2 Related Work}\label{related-work}

\subsection{2.1 Sign language dictionaries and form-based
access}\label{sign-language-dictionaries-and-form-based-access}

Chinese Sign Language exhibits regional variation
\citep{cslvariation2020}. China's National Common Sign Language program
provides a standardized lexical reference \citep{standard2018}, which
does not cover every sign used across communities.

Signbank supports sign-language dictionary organization and publication
\citep{signbank2018}. Reverse-access research includes a user-powered
American Sign Language dictionary \citep{aslsearch2015}, sign-lexicon
search \citep{signsearch2021}, and reduced feature sets for dictionary
lookup \citep{reduced2025}. Hybrid search supplements recognition with
user input \citep{hybrid2022}.

SignTrace complements these approaches with free-text movement
descriptions. Multiple query views accommodate incomplete or ambiguous
wording, and the returned dictionary entries let users inspect the
candidates.

\subsection{2.2 Reverse dictionaries and sign-video
retrieval}\label{reverse-dictionaries-and-sign-video-retrieval}

Written-language reverse dictionaries retrieve words from descriptions
of their meanings. WantWords demonstrates this information need in
English, Chinese, and cross-lingual lookup \citep{wantwords2020}. The
present task is related, but the description concerns visible form
rather than a definition. For example, a thumb moving near the nose
supplies evidence about an action; it does not directly define the
sign's meaning.

Duarte et al.~use cross-modal representations for sentence-to-video
retrieval with the How2Sign dataset and the SPOT-ALIGN framework
\citep{textsign2022}. SignTrace instead retrieves discrete dictionary
entries from text; it does not perform video recognition or continuous
sign-language translation.

\subsection{2.3 Query expansion, fusion, and LLM
reranking}\label{query-expansion-fusion-and-llm-reranking}

BM25 is a lexical retrieval model that matches query and document terms
\citep{bm252009}. Query2doc uses LLM-generated document-like text to
expand a query and reduce vocabulary mismatch \citep{query2doc2023}.
SignTrace applies this idea to the language of sign-dictionary movement
descriptions. The rewritten description supplements the unmodified query
and the structured restatement. The current controls examine the
contribution of the rewrite channel; they do not isolate writing style
from other changes introduced by rewriting.

Reciprocal rank fusion (RRF) combines the positions assigned by multiple
retrievers \citep{rrf2009}. Its behavior depends on the input rankings
and smoothing constant \citep{fusion2024}. We examine head weighting and
the contributions of individual channels while holding the recorded
query representations fixed.

LLMs can reorder retrieved candidates by relevance \citep{rankgpt2023}.
Their predictions can depend on candidate presentation order, motivating
work on permutation self-consistency \citep{permutation2024}. Our
reranker requests an action-match score for each candidate in a shared
prompt. The current evaluation fixes the reranking model and its
20-candidate window. Each query is processed once; candidate-order
sensitivity and repeated-call stability are not assessed.

\section{3 Task and Dictionary
Resource}\label{task-and-dictionary-resource}

\subsection{3.1 Description-to-entry
retrieval}\label{description-to-entry-retrieval}

Let \(\mathcal{D}=\{d_1,\ldots,d_N\}\) be a dictionary whose entries
associate a sign form with one or more written meanings. A query \(q\)
describes the observed movement in ordinary language. The system returns
a ranking \(\pi(q)\) of at most \(K\) entries. In this study,
\(N=6{,}699\) and \(K=60\).

An observation may come from a conversation, lesson, illustration, or
video; the system receives its textual description and retrieves
existing entries. The benchmark is closed-world: each query has a
designated target in the dictionary.

\subsection{3.2 Four layers of document
representation}\label{four-layers-of-document-representation}

The local corpus derives from the four-volume National Common Chinese
Sign Language Dictionary, published by Huaxia Publishing House in 2019
\citep{dictionary2019book, dictionarycorpus}. It stores 6,699
illustrated entry records and 8,687 listed meaning records. These counts
refer to the extracted corpus; each retrieved entry can have several
meanings.

Each indexed document combines four layers. The first is the source
action description. The second is a free-text visual re-description
generated from the illustration and associated description. The third
contains structured summaries of handshape, location, orientation,
contact, movement, two-hand relation, and other visible properties,
together with short visual tokens. The fourth is the list of written
senses. The first layer preserves the dictionary wording; the generated
layers make visually relevant details available in several formulations.

\subsection{3.3 Dictionary annotation}\label{dictionary-annotation}

Dictionary enrichment was completed before retrieval evaluation.
Structural annotations cover all 6,699 entries, and their metadata
record GPT-5.5-family model identifiers. Query processing uses DeepSeek
and Gemini. Free-text visual descriptions also cover all entries,
although their model identifiers and full generation prompts are
unavailable. The supplementary material documents the available
annotation metadata. No independent estimate of annotation accuracy is
available.

\section{4 SignTrace Architecture}\label{signtrace-architecture}

\subsection{4.1 Overview}\label{overview}

SignTrace enriches the dictionary before processing queries (Figure 1).
For each query, it extracts action features and rewrites the description
in parallel. Seven retrieval channels find candidate entries, and
head-weighted fusion selects a pool of 60. The reranker then scores the
first 20 candidates by how closely their movements match the query.

We use DeepSeek to understand and rewrite movement descriptions, and
Gemini to rerank candidate entries.

\begin{figure}
\centering
\includegraphics[width=1\linewidth,height=\textheight,keepaspectratio,alt={The SignTrace architecture: dictionary enrichment, parallel query transformations, seven-channel candidate retrieval, and action-based reranking. The dashed path supplies the extracted observation directly to the reranker.}]{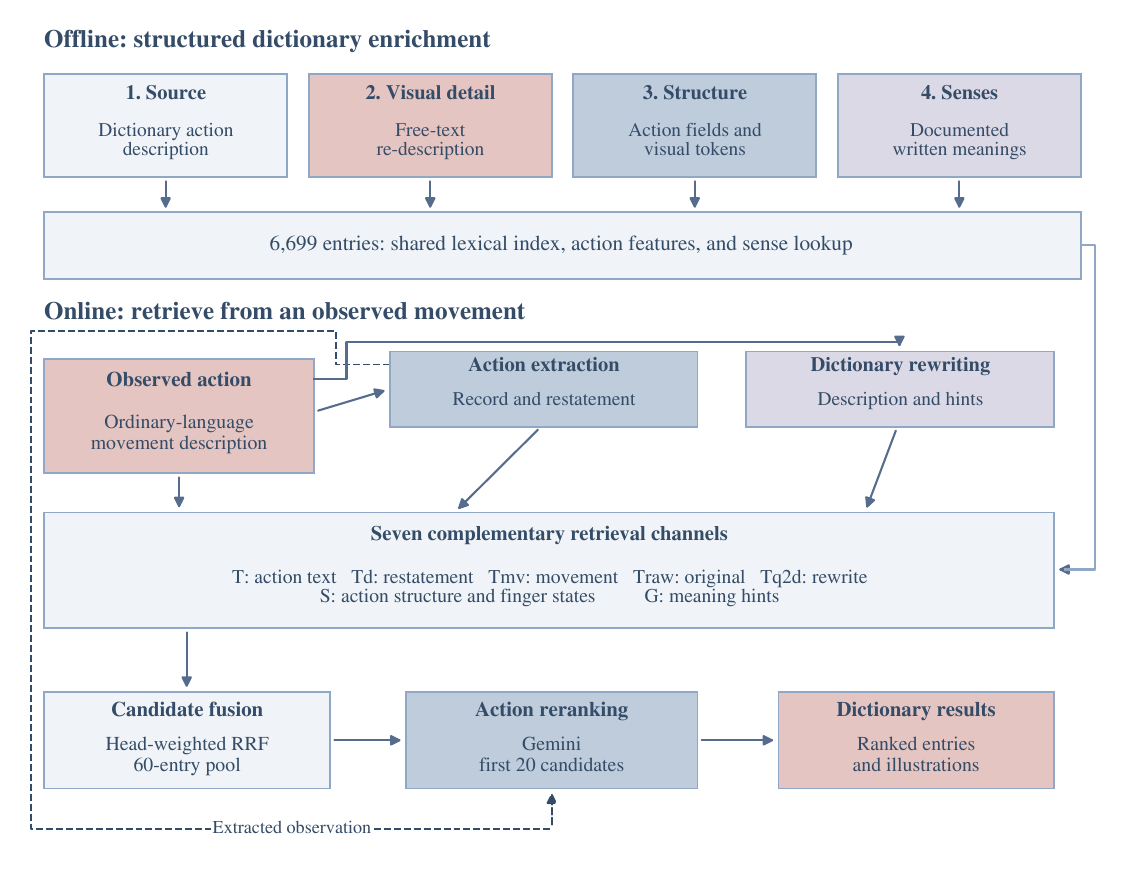}
\caption{The SignTrace architecture: dictionary enrichment, parallel
query transformations, seven-channel candidate retrieval, and
action-based reranking. The dashed path supplies the extracted
observation directly to the reranker.}\label{fig:architecture}
\end{figure}

\subsection{4.2 Structured action
extraction}\label{structured-action-extraction}

DeepSeek extracts the number of hands, handshape, finger states,
orientation, location, contact, movement, two-hand relation, and action
sequence from the input. It also restates the description, identifies
uncertain details, and suggests possible meanings in order of
likelihood. The prompt requires unspecified properties to remain unknown
and separates observed actions from guesses about meaning.

Selected action fields support broad lexical matching. The restatement
gives a compact description, while movement fields supply movement terms
and the described steps. BM25 does not explicitly align temporal
sequences. Discrete properties support structural comparison, while
retaining the original wording preserves expressions that
transformations may lose (Figure 2).

\begin{figure}
\centering
\includegraphics[width=1\linewidth,height=\textheight,keepaspectratio,alt={Complementary retrieval representations for example query 81. The input and selected fields are translated from the recorded Chinese query and extraction. Hand count, fingertip state, movement, location, and contact support complementary query representations; the unreported orientation remains unspecified.}]{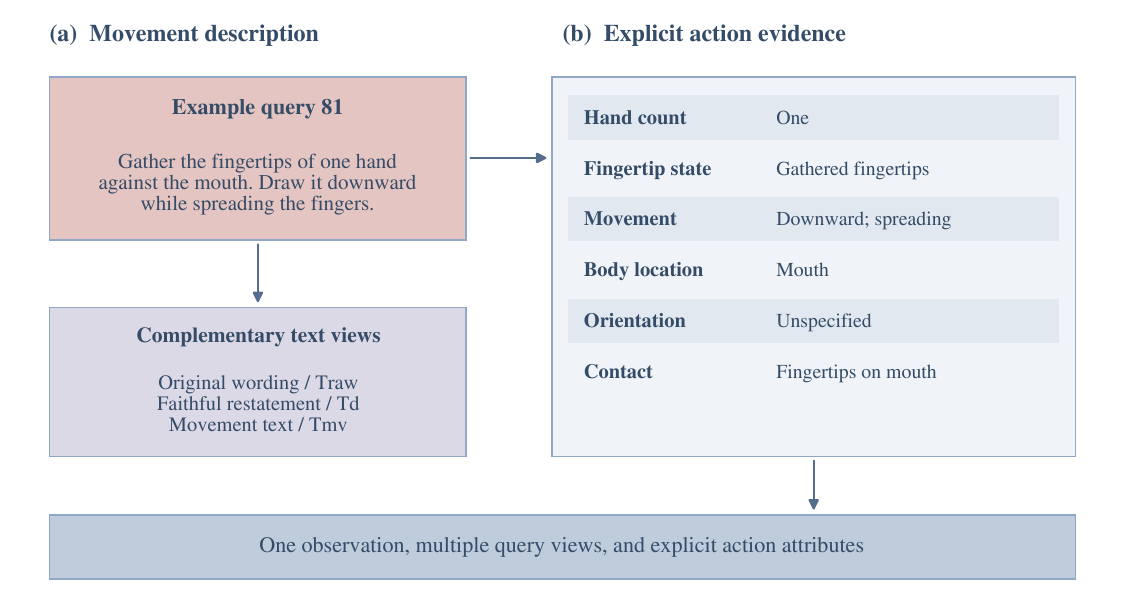}
\caption{Complementary retrieval representations for example query 81.
The input and selected fields are translated from the recorded Chinese
query and extraction. Hand count, fingertip state, movement, location,
and contact support complementary query representations; the unreported
orientation remains unspecified.}\label{fig:representation}
\end{figure}

\subsection{4.3 Parallel dictionary-style
rewriting}\label{parallel-dictionary-style-rewriting}

In parallel, DeepSeek rewrites the observation in the style of a
dictionary movement description, inspired by query2doc
\citep{query2doc2023}. The prompt requests 60-120 Chinese characters
organized around handshape, orientation, location, and movement, without
adding unspecified details. It also asks for five possible meanings on a
separate line.

The rewrite supplies a separate text channel, and its meaning guesses
augment the meaning-hint channel. Reranking still receives the original
extraction, so additional rewrite guesses do not become new action
observations.

\subsection{4.4 Five textual retrieval
channels}\label{five-textual-retrieval-channels}

All five textual views search a shared BM25 index \citep{bm252009}.
Chinese sequences are tokenized into individual characters and adjacent
character pairs. Latin-letter and digit runs remain whole. Query tokens
are deduplicated; their frequency in the query is not used. For
query-view token set \(Q_j\) and document \(d\),

\[
\operatorname{BM25}(Q_j,d)=\sum_{t\in Q_j}\log\!\left(1+\frac{N-n_t+0.5}{n_t+0.5}\right)
\frac{f(t,d)(k_1+1)}{f(t,d)+k_1(1-b+b|d|/\overline{L})}.
\]

Here, \(n_t\) is the number of documents containing token \(t\),
\(f(t,d)\) is its term frequency within document \(d\), and
\(\overline{L}\) is mean document length. The implementation uses
\(k_1=1.5\) and \(b=0.75\).

The channels are \textbf{T}, the textual action fields of the structured
record; \textbf{Td}, its faithful restatement; \textbf{Tmv}, movement
keywords and steps; \textbf{Traw}, the unmodified query; and
\textbf{Tq2d}, the dictionary-style rewrite. T and Td retain every
positive-scoring entry for fusion. Tmv, Traw, and Tq2d contribute only
their first 20 entries.

\subsection{4.5 Structural and meaning-hint
channels}\label{structural-and-meaning-hint-channels}

The \textbf{S} channel compares supported action attributes. Its
components cover hand count, individual finger states, movement
families, body regions, orientation, and contact. It combines component
scores through a weighted mean. The finger-state component has weight 2;
the other components have weight 1. Finger-state, region, and
orientation comparisons abstain when the required evidence cannot be
compared. Hand-count and contact components use the stored categorical
values.

Finger matching distinguishes extension and flexion, handles
intermediate states conservatively, and gives greater weight to rarer
finger-state features. For two hands, the better of the normal and
exchanged assignments is retained. Orientation comparison tolerates
left-right mirroring. These rules apply to every query and accommodate
ambiguity in the observer's viewpoint. Movement expressions are
normalized to shared motion families before matching. S retains positive
scores without a fixed rank cutoff.

The \textbf{G} channel compares ordered meaning guesses with entry
senses. Exact matches receive more weight than partial matches, and
earlier guesses receive more weight than later ones. One-character
senses do not support partial matching. Only the first ten entries
contribute votes. This channel treats a plausible meaning as a retrieval
hint. The later action-match assessment is instructed to judge the
described form rather than reward agreement with a guessed word.

\subsection{4.6 Candidate fusion}\label{candidate-fusion}

Let \(r_j(d)\) be the one-based position of entry \(d\) in channel \(j\)
after that channel's cutoff. Let \(\mathcal{J}(d)\) contain the channels
that retained the entry. The fusion score is

\[
F(q,d)=\sum_{j\in\mathcal{J}(d)}\frac{1+w\,\mathbf{1}[r_j(d)\leq m]}{c+r_j(d)}.
\]

We use \(c=300\), \(m=3\), and \(w=1.5\). Each channel's first three
entries therefore receive 2.5 times the ordinary RRF contribution.
Entries absent from a list receive no vote, and entry numbers break
ties. The first 60 entries form the candidate pool.

The head bonus increases the influence of a channel's strongest matches
relative to weak support distributed across long lists. The structural
channel, in particular, can return many partially compatible forms. The
offline controls compare the same channel outputs with and without the
head bonus and after individual channel removals.

Figure 3 illustrates query 490, whose target entry is ``predicate''.
Five channels contribute votes for this entry. It ranks thirteenth after
fusion and first after Gemini reranking. This example illustrates how a
candidate with distributed support can reach the comparison stage even
when no channel places it first.

\begin{figure}
\centering
\includegraphics[width=1\linewidth,height=\textheight,keepaspectratio,alt={Recorded target votes for query 490. Bars show contributions after channel cutoffs; an absent vote is zero. The target moves from recall rank 13 to final rank 1.}]{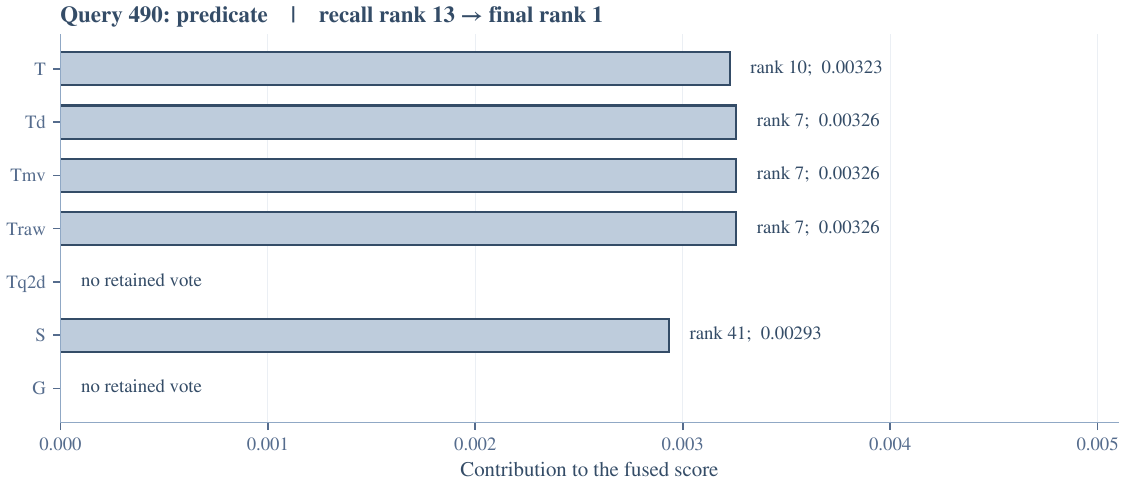}
\caption{Recorded target votes for query 490. Bars show contributions
after channel cutoffs; an absent vote is zero. The target moves from
recall rank 13 to final rank 1.}\label{fig:fusionexample}
\end{figure}

\FloatBarrier

\subsection{4.7 Action-grounded
reranking}\label{action-grounded-reranking}

The reranker receives the extracted observation and the first \(h=20\)
candidates in one prompt. Each candidate contributes action descriptions
and structured summaries truncated to 200 characters after trimming. The
explicit sense list is omitted, but the retained prose can still contain
semantic explanations. The prompt requests scores from 0 to 100 for form
compatibility, emphasizing handshape and individual fingers, and
instructs the model not to use meaning guesses as scoring evidence.
These instructions specify the intended scoring rule; they do not
establish that all semantic influence has been excluded.

Candidates are sorted by score; ties preserve recall order, and unscored
candidates follow scored ones. Entries outside the window retain their
recall order. Reranking can promote a target within the window but
cannot recover one outside it into the leading positions.

The evaluated system uses Gemini's ordering directly, without blending
it with the initial ranks. The 60-entry pool is retained. All reranking
system messages retain an image-workflow instruction that conditions
movement scoring on confirmed multi-frame observation. An uncertainty
heuristic can add a static-image caution that instructs the reranker to
ignore movement. It is triggered by uncertainty terms in the movement
field, or by combined movement and uncertainty terms in the uncertainty
list. All queries in this study are textual; the inherited caution can
therefore be inappropriate for a described action. A malformed
extraction response separately invokes an original-query fallback.
Section 6.5 reports the occurrence and outcomes of these behaviors.

\section{5 Experimental Design}\label{experimental-design}

\subsection{5.1 Evaluation dataset}\label{evaluation-dataset}

The evaluation dataset comprises 500 Chinese movement-description
queries, each paired with a distinct target dictionary entry. The
queries are derived from dictionary action descriptions and annotations.
Query texts and target entries were fixed before evaluation, and each
query was processed once using the configuration in Section 5.3. Table 1
summarizes the dataset.

\FloatBarrier
\Needspace{13\baselineskip}

\begin{longtable}[]{@{}
  >{\raggedright\arraybackslash}p{(\linewidth - 2\tabcolsep) * \real{0.4300}}
  >{\raggedright\arraybackslash}p{(\linewidth - 2\tabcolsep) * \real{0.5700}}@{}}
\caption{\textbf{Table 1.} Characteristics of the evaluation
dataset.}\tabularnewline
\toprule\noalign{}
\begin{minipage}[b]{\linewidth}\raggedright
Property
\end{minipage} & \begin{minipage}[b]{\linewidth}\raggedright
Value
\end{minipage} \\
\midrule\noalign{}
\endfirsthead
\toprule\noalign{}
\begin{minipage}[b]{\linewidth}\raggedright
Property
\end{minipage} & \begin{minipage}[b]{\linewidth}\raggedright
Value
\end{minipage} \\
\midrule\noalign{}
\endhead
\bottomrule\noalign{}
\endlastfoot
Query source & Dictionary action descriptions and annotations \\
Queries and target entries & 500 / 500 \\
Query identifiers & 1--500 \\
Description length, including punctuation & 13--64 characters; median
35 \\
Attempted / scored / excluded & 500 / 500 / 0 \\
\end{longtable}

Descriptions range from 13 to 64 Unicode characters, including
punctuation, with a median of 35. All method comparisons use the
complete dataset; descriptive subgroup analyses are identified
separately. The dataset includes 200 queries whose targets had
previously been selected for retrieval within a 60-entry candidate pool.
This introduces selection bias, as discussed in Section 8.2.

\FloatBarrier
\Needspace{16\baselineskip}

\begin{longtable}[]{@{}
  >{\raggedright\arraybackslash}p{(\linewidth - 6\tabcolsep) * \real{0.0550}}
  >{\raggedright\arraybackslash}p{(\linewidth - 6\tabcolsep) * \real{0.6400}}
  >{\raggedright\arraybackslash}p{(\linewidth - 6\tabcolsep) * \real{0.2050}}
  >{\raggedright\arraybackslash}p{(\linewidth - 6\tabcolsep) * \real{0.1000}}@{}}
\caption{\textbf{Table 2.} Example evaluation queries and final target
ranks, including a retrieval error and a successful fallback case.
English descriptions and glosses translate the Chinese queries and
reference meanings. Only movement descriptions are submitted as
input.}\tabularnewline
\toprule\noalign{}
\begin{minipage}[b]{\linewidth}\raggedright
ID
\end{minipage} & \begin{minipage}[b]{\linewidth}\raggedright
Movement description
\end{minipage} & \begin{minipage}[b]{\linewidth}\raggedright
Target meaning
\end{minipage} & \begin{minipage}[b]{\linewidth}\raggedright
Rank
\end{minipage} \\
\midrule\noalign{}
\endfirsthead
\toprule\noalign{}
\begin{minipage}[b]{\linewidth}\raggedright
ID
\end{minipage} & \begin{minipage}[b]{\linewidth}\raggedright
Movement description
\end{minipage} & \begin{minipage}[b]{\linewidth}\raggedright
Target meaning
\end{minipage} & \begin{minipage}[b]{\linewidth}\raggedright
Rank
\end{minipage} \\
\midrule\noalign{}
\endhead
\bottomrule\noalign{}
\endlastfoot
1 & Hold the left thumb upright. Stroke its back with the right hand,
then hold the right hand across its front and lower both hands together.
& Care for; protect & 1 \\
81 & Gather the fingertips of one hand against the mouth, then draw the
hand downward while spreading the fingers. & Experience; try & 1 \\
201 & Raise the left thumb. Move the curved fingers of the right hand
from the left shoulder onto the thumb. Then form a ring with the left
thumb and index finger and flick the bent right index and middle fingers
outward twice at its opening. & Supported care & 1 \\
343 & Hold the left hand open and upright, palm outward. Bend the right
index and middle fingers and tap their tips against the left palm. &
Numeral; classifier & 13 \\
489 & Point both hands downward and rotate the wrists in front of the
body. Then curve the fingers slightly and cup one hand over each ear. &
Audiometry & 1 \\
490 & Hold the index, middle, and ring fingers apart and upright,
pressing the bent little finger with the thumb. Then hold the index
finger horizontally in front of the mouth and rotate it back and forth
twice. & Predicate & 1 \\
\end{longtable}

\subsection{5.2 Evaluation metrics and statistical
analysis}\label{evaluation-metrics-and-statistical-analysis}

Retrieval is evaluated against the target entry assigned to each query.
The reference targets remain fixed throughout evaluation; alternative
entries with similar meanings are not counted as matches. The target
identifiers are not provided to the models, and candidate sense lists
are omitted as separate fields. However, candidate action descriptions
and generated meaning hypotheses may retain semantic cues. The
evaluation therefore measures dictionary lookup performance under these
input conditions.

If \(r_i\) is the designated target's rank and an absent target has rank
\(\infty\), then

\[
\operatorname{Hit@}k=\frac{1}{n}\sum_{i=1}^{n}\mathbf{1}[r_i\leq k],\qquad
\operatorname{MRR}=\frac{1}{n}\sum_{i=1}^{n}\frac{1}{r_i},
\]

where \(1/\infty=0\). Every ranking is truncated to 60 for evaluation,
including the lexical controls. Hit@1 measures whether the target is
first. Hit@9 evaluates a nine-entry shortlist, while Hit@3 and Hit@5
characterize shorter result lists. Hit@20 measures access to the
reranker, and Pool@60 measures candidate-pool coverage. No attempted
query is excluded because of a parsing or processing failure.

We report counts, percentages, and 95\% Wilson intervals for
proportions. Four paired comparisons on all 500 queries examine
reranking at cutoffs 1 and 9 and seven-channel versus raw-query BM25
retrieval at cutoffs 20 and 60. Exact two-sided tests use the binomial
distribution of wins among discordant pairs, with Holm correction across
these four comparisons. Conservative 95\% intervals for paired accuracy
differences are constructed from simultaneous win/loss probability
intervals (Appendix A). These analyses are exploratory. Their
uncertainty calculations treat the evaluation queries as sampling units;
they do not correct for dictionary-derived wording, prior selection of
benchmark queries, or variation across repeated API runs.

\subsection{5.3 Experimental
configuration}\label{experimental-configuration}

DeepSeek performs action extraction and dictionary-style rewriting in
parallel, and Gemini reranks the first 20 of the 60 fused candidates.
The retrieval parameters are specified in Section 4. Evaluation was
conducted on September 13, 2026, with application-level model-output
caches disabled and at most six concurrent queries. Queries were
processed in a randomized order, and the total batch duration was 19
minutes 29 seconds.

All 500 queries received two DeepSeek calls and one Gemini call, for
1,500 HTTP requests. There were no HTTP errors, retries, missing
rewrites, or reranking failures. Three extraction responses could not be
parsed and used the original-query fallback. These cases remain in all
reported results. Gemini was used for every reranking call, and all
calls succeeded. Appendix B records requested and returned model
identifiers, distinguishing endpoint labels from immutable model
releases.

Each query's processing time is measured from the start of preparation
to the completion of reranking and local result recording. It includes
parallel model preparation, local retrieval, and reranking, but excludes
waiting for a query worker and browser or public-service network
overhead. We report this processing time separately from batch duration
and model token usage.

\subsection{5.4 Baselines and component
analyses}\label{baselines-and-component-analyses}

Raw-query BM25 searches the same enriched dictionary using the
unmodified input. We compare it with seven-channel retrieval before
reranking and the recorded final SignTrace ranking. This separates
first-stage quality from the measured effect of reranking on its actual
candidate lists. Gemini was not run on the BM25 lists, so this study
does not establish a final-ranking advantage over BM25 followed by the
same reranker.

Two sets of offline controls examine the first stage. A
dictionary-enrichment analysis runs raw-query BM25 on the source action
description, then cumulatively adds visual re-description, structured
summaries and visual tokens, and sense strings. A channel control
removes each of T, Td, Tmv, Traw, Tq2d, S, and G in turn; a further
control removes the RRF head bonus. Removing Tq2d retains the rewrite's
generated meaning guesses in G, so it isolates the rewrite text channel
rather than all rewriting-related evidence.

Component comparisons hold the model-generated query representations
fixed and modify only indexing or rank fusion. The baseline
configuration reproduces the original 60-entry retrieval order for every
query. Alternative candidate lists are evaluated at the retrieval stage
without additional model calls or reranking.

\section{6 Results}\label{results}

\subsection{6.1 Retrieval performance}\label{retrieval-performance}

SignTrace achieves 94.0\% Hit@1 (470/500; 95\% Wilson confidence
interval, 91.6\%--95.8\%) and 97.4\% Hit@9 (487/500; 95\% confidence
interval, 95.6\%--98.5\%). Candidate-pool coverage is 99.4\% (497/500).
Table 3 summarizes the retrieval results.

\FloatBarrier
\Needspace{12\baselineskip}

\begin{longtable}[]{@{}
  >{\raggedright\arraybackslash}p{(\linewidth - 6\tabcolsep) * \real{0.2400}}
  >{\raggedright\arraybackslash}p{(\linewidth - 6\tabcolsep) * \real{0.1800}}
  >{\raggedright\arraybackslash}p{(\linewidth - 6\tabcolsep) * \real{0.2000}}
  >{\raggedright\arraybackslash}p{(\linewidth - 6\tabcolsep) * \real{0.3800}}@{}}
\caption{\textbf{Table 3.} Final performance on the complete
dictionary-derived benchmark. All proportions use a denominator of 500;
no attempted query is excluded.}\tabularnewline
\toprule\noalign{}
\begin{minipage}[b]{\linewidth}\raggedright
Metric
\end{minipage} & \begin{minipage}[b]{\linewidth}\raggedright
Hits / total
\end{minipage} & \begin{minipage}[b]{\linewidth}\raggedright
Rate (\%)
\end{minipage} & \begin{minipage}[b]{\linewidth}\raggedright
95\% Wilson CI (\%)
\end{minipage} \\
\midrule\noalign{}
\endfirsthead
\toprule\noalign{}
\begin{minipage}[b]{\linewidth}\raggedright
Metric
\end{minipage} & \begin{minipage}[b]{\linewidth}\raggedright
Hits / total
\end{minipage} & \begin{minipage}[b]{\linewidth}\raggedright
Rate (\%)
\end{minipage} & \begin{minipage}[b]{\linewidth}\raggedright
95\% Wilson CI (\%)
\end{minipage} \\
\midrule\noalign{}
\endhead
\bottomrule\noalign{}
\endlastfoot
Hit@1 & 470/500 & 94.0 & 91.6--95.8 \\
Hit@9 & 487/500 & 97.4 & 95.6--98.5 \\
Pool@60 & 497/500 & 99.4 & 98.3--99.8 \\
\end{longtable}

The mean reciprocal rank is 0.9540 at a maximum ranking depth of 60.
Figure 4 compares raw-query BM25, seven-channel retrieval, and the final
ranking across rank cutoffs. Appendix C reports the corresponding
counts.

\begin{figure}
\centering
\includegraphics[width=1\linewidth,height=\textheight,keepaspectratio,alt={Retrieval hit rates on the evaluation dataset (n = 500). All curves use the same maximum depth of 60. Line styles and markers distinguish raw-query BM25, seven-channel recall, and the final ranking. Vertical guides mark the nine-result cutoff and the 20-candidate reranking window.}]{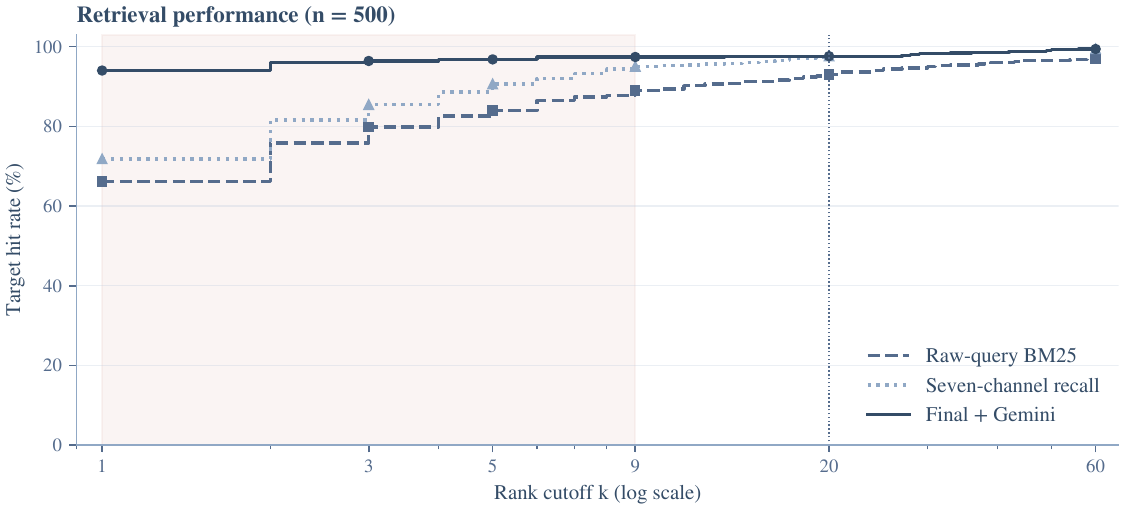}
\caption{Retrieval hit rates on the evaluation dataset (n = 500). All
curves use the same maximum depth of 60. Line styles and markers
distinguish raw-query BM25, seven-channel recall, and the final ranking.
Vertical guides mark the nine-result cutoff and the 20-candidate
reranking window.}\label{fig:hitcurves}
\end{figure}

\subsection{6.2 Effect of reranking}\label{effect-of-reranking}

Across the evaluation dataset, raw-query BM25 over enriched entries
places 331 targets first, and seven-channel recall places 359 first.
Gemini raises this to 470, a gain of 22.2 percentage points over the
same candidate lists before reranking (Table 4). Of the changed
first-result outcomes, 114 improve and three worsen. The conservative
paired interval for the gain is 16.8--27.2 points, and the Holm-adjusted
exact p-value is \(1.29\times10^{-29}\) (Table 5).

\FloatBarrier
\Needspace{12\baselineskip}

\begin{longtable}[]{@{}
  >{\raggedright\arraybackslash}p{(\linewidth - 10\tabcolsep) * \real{0.3400}}
  >{\raggedright\arraybackslash}p{(\linewidth - 10\tabcolsep) * \real{0.1200}}
  >{\raggedright\arraybackslash}p{(\linewidth - 10\tabcolsep) * \real{0.1200}}
  >{\raggedright\arraybackslash}p{(\linewidth - 10\tabcolsep) * \real{0.1200}}
  >{\raggedright\arraybackslash}p{(\linewidth - 10\tabcolsep) * \real{0.1600}}
  >{\raggedright\arraybackslash}p{(\linewidth - 10\tabcolsep) * \real{0.1400}}@{}}
\caption{\textbf{Table 4.} Stage comparison on the same 500 inputs and
enriched dictionary. All hit columns are counts. Only the SignTrace
candidate lists received Gemini reranking.}\tabularnewline
\toprule\noalign{}
\begin{minipage}[b]{\linewidth}\raggedright
Method
\end{minipage} & \begin{minipage}[b]{\linewidth}\raggedright
Hit@1
\end{minipage} & \begin{minipage}[b]{\linewidth}\raggedright
Hit@9
\end{minipage} & \begin{minipage}[b]{\linewidth}\raggedright
Hit@20
\end{minipage} & \begin{minipage}[b]{\linewidth}\raggedright
Pool@60
\end{minipage} & \begin{minipage}[b]{\linewidth}\raggedright
MRR
\end{minipage} \\
\midrule\noalign{}
\endfirsthead
\toprule\noalign{}
\begin{minipage}[b]{\linewidth}\raggedright
Method
\end{minipage} & \begin{minipage}[b]{\linewidth}\raggedright
Hit@1
\end{minipage} & \begin{minipage}[b]{\linewidth}\raggedright
Hit@9
\end{minipage} & \begin{minipage}[b]{\linewidth}\raggedright
Hit@20
\end{minipage} & \begin{minipage}[b]{\linewidth}\raggedright
Pool@60
\end{minipage} & \begin{minipage}[b]{\linewidth}\raggedright
MRR
\end{minipage} \\
\midrule\noalign{}
\endhead
\bottomrule\noalign{}
\endlastfoot
Raw-query BM25 & 331 & 445 & 465 & 485 & 0.7446 \\
Seven-channel recall & 359 & 475 & 488 & 497 & 0.8002 \\
SignTrace + Gemini & 470 & 487 & 488 & 497 & 0.9540 \\
\end{longtable}

The reranker has a smaller effect at the nine-result cutoff: 12 targets
enter the first nine with no losses, moving from 475 to 487 hits (2.4
percentage points; adjusted p-value \(9.77\times10^{-4}\)). Relative to
enriched raw-query BM25, seven-channel recall adds 23 net targets to the
first 20, from 465 to 488, with 24 wins and one loss (adjusted p-value
\(4.65\times10^{-6}\)). Pool coverage rises from 485 to 497, with 12
wins and no losses (adjusted p-value \(9.77\times10^{-4}\)). These
paired gains concern the fixed dictionary-derived benchmark; they do not
remove its construction and selection biases.

\FloatBarrier
\Needspace{14\baselineskip}

\begin{longtable}[]{@{}
  >{\raggedright\arraybackslash}p{(\linewidth - 10\tabcolsep) * \real{0.2400}}
  >{\raggedright\arraybackslash}p{(\linewidth - 10\tabcolsep) * \real{0.0450}}
  >{\raggedright\arraybackslash}p{(\linewidth - 10\tabcolsep) * \real{0.2550}}
  >{\raggedright\arraybackslash}p{(\linewidth - 10\tabcolsep) * \real{0.1400}}
  >{\raggedright\arraybackslash}p{(\linewidth - 10\tabcolsep) * \real{0.1600}}
  >{\raggedright\arraybackslash}p{(\linewidth - 10\tabcolsep) * \real{0.1600}}@{}}
\caption{\textbf{Table 5.} Paired comparisons on the evaluation dataset
(n = 500). Gains are percentage points. Difference intervals are
conservative simultaneous-binomial intervals; exact p-values use
discordant pairs. Holm correction applies to the four displayed
comparisons.}\tabularnewline
\toprule\noalign{}
\begin{minipage}[b]{\linewidth}\raggedright
Comparison
\end{minipage} & \begin{minipage}[b]{\linewidth}\raggedright
k
\end{minipage} & \begin{minipage}[b]{\linewidth}\raggedright
Gain, pp {[}95\% CI{]}
\end{minipage} & \begin{minipage}[b]{\linewidth}\raggedright
Wins/losses
\end{minipage} & \begin{minipage}[b]{\linewidth}\raggedright
p
\end{minipage} & \begin{minipage}[b]{\linewidth}\raggedright
Holm p
\end{minipage} \\
\midrule\noalign{}
\endfirsthead
\toprule\noalign{}
\begin{minipage}[b]{\linewidth}\raggedright
Comparison
\end{minipage} & \begin{minipage}[b]{\linewidth}\raggedright
k
\end{minipage} & \begin{minipage}[b]{\linewidth}\raggedright
Gain, pp {[}95\% CI{]}
\end{minipage} & \begin{minipage}[b]{\linewidth}\raggedright
Wins/losses
\end{minipage} & \begin{minipage}[b]{\linewidth}\raggedright
p
\end{minipage} & \begin{minipage}[b]{\linewidth}\raggedright
Holm p
\end{minipage} \\
\midrule\noalign{}
\endhead
\bottomrule\noalign{}
\endlastfoot
Final vs.~recall & 1 & 22.2 {[}16.8, 27.2{]} & 114/3 & 3.21e-30 &
1.29e-29 \\
Final vs.~recall & 9 & 2.4 {[}0.3, 4.4{]} & 12/0 & 4.88e-04 &
9.77e-04 \\
Recall vs.~BM25 & 20 & 4.6 {[}1.6, 7.4{]} & 24/1 & 1.55e-06 &
4.65e-06 \\
Recall vs.~BM25 & 60 & 2.4 {[}0.3, 4.4{]} & 12/0 & 4.88e-04 &
9.77e-04 \\
\end{longtable}

Reranking leaves window and pool membership unchanged: Hit@20 is 488/500
and Pool@60 is 497/500 both before and after Gemini. Thus the large
first-result improvement reflects better ordering among available
candidates. It cannot compensate for a target that never reaches the
comparison window.

\subsection{6.3 Dictionary enrichment and channel
contributions}\label{dictionary-enrichment-and-channel-contributions}

Adding searchable action details improves retrieval in the
dictionary-enrichment analysis (Table 6). Raw-query BM25 retrieves 249
targets first using source descriptions. Adding visual re-descriptions
raises this to 265; adding structured summaries and visual tokens raises
it to 333. First-20 coverage rises from 403 to 419 and then 465.
Including sense strings produces 331 first-result hits and leaves
first-20 coverage at 465. These results show the benefit of enriched
text on this benchmark. Because the queries also derive from dictionary
annotations, this comparison cannot separate the effect of added action
information from that of similarity between query and index wording.

\FloatBarrier
\Needspace{13\baselineskip}

\begin{longtable}[]{@{}
  >{\raggedright\arraybackslash}p{(\linewidth - 8\tabcolsep) * \real{0.4000}}
  >{\raggedright\arraybackslash}p{(\linewidth - 8\tabcolsep) * \real{0.1500}}
  >{\raggedright\arraybackslash}p{(\linewidth - 8\tabcolsep) * \real{0.1500}}
  >{\raggedright\arraybackslash}p{(\linewidth - 8\tabcolsep) * \real{0.1500}}
  >{\raggedright\arraybackslash}p{(\linewidth - 8\tabcolsep) * \real{0.1500}}@{}}
\caption{\textbf{Table 6.} Raw-query BM25 with cumulative document
enrichment on the evaluation dataset (n = 500). Cells are first-stage
hit counts. The query itself is unchanged.}\tabularnewline
\toprule\noalign{}
\begin{minipage}[b]{\linewidth}\raggedright
Lexical document layers
\end{minipage} & \begin{minipage}[b]{\linewidth}\raggedright
Hit@1
\end{minipage} & \begin{minipage}[b]{\linewidth}\raggedright
Hit@9
\end{minipage} & \begin{minipage}[b]{\linewidth}\raggedright
Hit@20
\end{minipage} & \begin{minipage}[b]{\linewidth}\raggedright
Pool@60
\end{minipage} \\
\midrule\noalign{}
\endfirsthead
\toprule\noalign{}
\begin{minipage}[b]{\linewidth}\raggedright
Lexical document layers
\end{minipage} & \begin{minipage}[b]{\linewidth}\raggedright
Hit@1
\end{minipage} & \begin{minipage}[b]{\linewidth}\raggedright
Hit@9
\end{minipage} & \begin{minipage}[b]{\linewidth}\raggedright
Hit@20
\end{minipage} & \begin{minipage}[b]{\linewidth}\raggedright
Pool@60
\end{minipage} \\
\midrule\noalign{}
\endhead
\bottomrule\noalign{}
\endlastfoot
1: Source description & 249 & 369 & 403 & 434 \\
1+2: Visual re-description & 265 & 390 & 419 & 451 \\
1+2+3: Structured summaries & 333 & 446 & 465 & 485 \\
1+2+3+4: Sense strings & 331 & 445 & 465 & 485 \\
\end{longtable}

Removing the rewrite text channel lowers first-result hits from 359 to
324, first-nine hits from 475 to 459, and first-20 coverage from 488 to
479 (Table 7). This is the largest first-result reduction among
individual channel removals. The rewritten description therefore
contributes to retrieval on this benchmark. The comparison retains the
meaning hints generated alongside it.

\FloatBarrier
\Needspace{19\baselineskip}

\begin{longtable}[]{@{}
  >{\raggedright\arraybackslash}p{(\linewidth - 8\tabcolsep) * \real{0.4000}}
  >{\raggedright\arraybackslash}p{(\linewidth - 8\tabcolsep) * \real{0.1500}}
  >{\raggedright\arraybackslash}p{(\linewidth - 8\tabcolsep) * \real{0.1500}}
  >{\raggedright\arraybackslash}p{(\linewidth - 8\tabcolsep) * \real{0.1500}}
  >{\raggedright\arraybackslash}p{(\linewidth - 8\tabcolsep) * \real{0.1500}}@{}}
\caption{\textbf{Table 7.} Retrieval-channel and fusion ablations on the
evaluation dataset (n = 500). Cells are hit counts. Removing Tq2d
retains its meaning guesses in G. All configurations share the same
model outputs; changed candidate lists are not reranked.}\tabularnewline
\toprule\noalign{}
\begin{minipage}[b]{\linewidth}\raggedright
First-stage configuration
\end{minipage} & \begin{minipage}[b]{\linewidth}\raggedright
Hit@1
\end{minipage} & \begin{minipage}[b]{\linewidth}\raggedright
Hit@9
\end{minipage} & \begin{minipage}[b]{\linewidth}\raggedright
Hit@20
\end{minipage} & \begin{minipage}[b]{\linewidth}\raggedright
Pool@60
\end{minipage} \\
\midrule\noalign{}
\endfirsthead
\toprule\noalign{}
\begin{minipage}[b]{\linewidth}\raggedright
First-stage configuration
\end{minipage} & \begin{minipage}[b]{\linewidth}\raggedright
Hit@1
\end{minipage} & \begin{minipage}[b]{\linewidth}\raggedright
Hit@9
\end{minipage} & \begin{minipage}[b]{\linewidth}\raggedright
Hit@20
\end{minipage} & \begin{minipage}[b]{\linewidth}\raggedright
Pool@60
\end{minipage} \\
\midrule\noalign{}
\endhead
\bottomrule\noalign{}
\endlastfoot
All seven channels & 359 & 475 & 488 & 497 \\
Without T & 353 & 469 & 481 & 494 \\
Without Td & 357 & 474 & 484 & 495 \\
Without Tmv & 359 & 472 & 488 & 498 \\
Without Traw & 360 & 474 & 485 & 498 \\
Without Tq2d & 324 & 459 & 479 & 494 \\
Without S & 365 & 471 & 484 & 497 \\
Without G & 360 & 474 & 487 & 498 \\
Flat RRF & 271 & 467 & 487 & 497 \\
\end{longtable}

The channel ablations reveal differences between ranking accuracy and
candidate coverage. Removing S increases first-result hits to 365 but
lowers first-20 coverage to 484. Removing G increases first-result hits
to 360 while reducing first-20 coverage to 487. Removing the head bonus
decreases first-result hits from 359 to 271, while first-20 coverage
decreases by one target and pool coverage remains unchanged. The
selected fusion is therefore not optimal on every first-stage metric.
Figure 5 summarizes enrichment and the channel-removal effects on
candidate availability; final-ranking comparisons of alternative
candidate lists require additional reranking experiments.

\begin{figure}
\centering
\includegraphics[width=1\linewidth,height=\textheight,keepaspectratio,alt={Component analyses on the evaluation dataset (n = 500). Left: first-result accuracy of unmodified-query BM25 as entry text is enriched. Right: changes in first-20 hit counts after removing individual channels, relative to 488 hits with all seven channels. The shared query representations are fixed, and no altered candidate list is reranked.}]{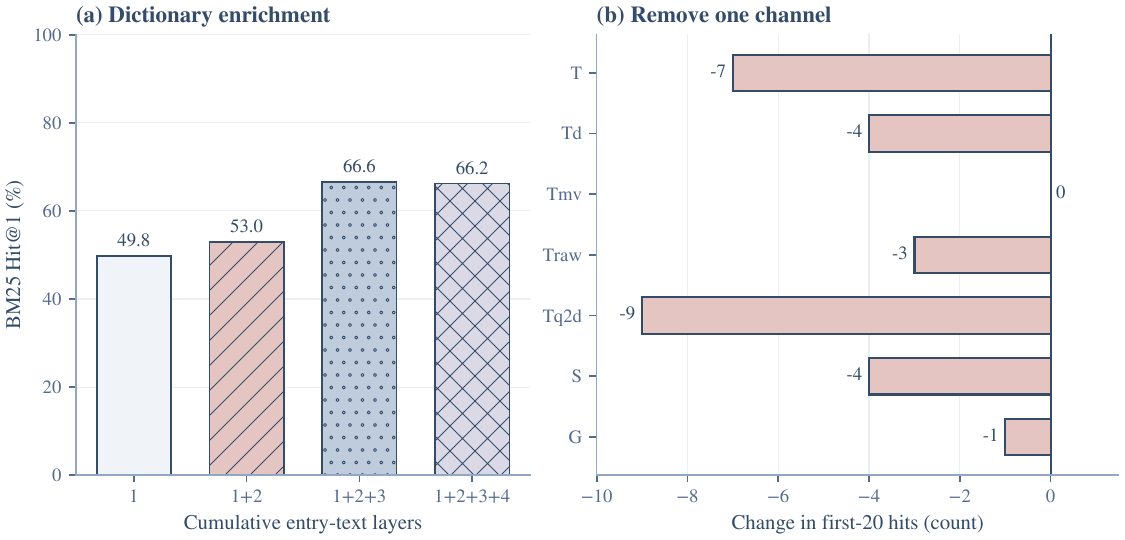}
\caption{Component analyses on the evaluation dataset (n = 500). Left:
first-result accuracy of unmodified-query BM25 as entry text is
enriched. Right: changes in first-20 hit counts after removing
individual channels, relative to 488 hits with all seven channels. The
shared query representations are fixed, and no altered candidate list is
reranked.}\label{fig:enrichment}
\end{figure}

\subsection{6.4 Error analysis and query
length}\label{error-analysis-and-query-length}

Thirteen targets fall outside the first nine final results in the
evaluation dataset. Nine remain at recall positions 21--60 and three are
absent from the pool. The remaining target, query 343
(numeral/classifier), enters the window at rank 16 and finishes at rank
13. Thus 12 of the 13 failures at this cutoff are constrained by
candidate selection. Appendix C lists every failed target and both
ranks, identified by query index.

This bottleneck depends on the metric. At the first-result cutoff, 30
queries fail: 18 have targets within the reranking window, and 12 are
outside it. Improving action discrimination within the window is
therefore also relevant to first-result accuracy. Expanding the window
would expose some missed targets to Gemini, but this run does not
measure its accuracy, latency, or cost at larger windows.

Across the complete benchmark, 40/45 descriptions of 13--24 characters
produce a first-result hit, compared with 286/307 of 25--40 characters
and 144/148 of 41--64 characters. These three post hoc strata partition
all 500 queries. They contain different signs and different action
complexity, so the pattern does not establish a causal effect of length.
Appendix C reports counts and intervals.

\subsection{6.5 Processing time and
reliability}\label{processing-time-and-reliability}

The median query-processing time is 13.37 seconds, with 90th and 95th
percentiles of 17.14 and 19.36 seconds (Table 8). Parallel preparation
and local recall have a median of 2.69 seconds; Gemini reranking has a
median of 10.68 seconds. These component medians need not sum to the
median of per-query totals. Figure 6 shows the complete distributions
under six concurrent queries.

\FloatBarrier
\Needspace{13\baselineskip}

\begin{longtable}[]{@{}
  >{\raggedright\arraybackslash}p{(\linewidth - 6\tabcolsep) * \real{0.4000}}
  >{\raggedright\arraybackslash}p{(\linewidth - 6\tabcolsep) * \real{0.2000}}
  >{\raggedright\arraybackslash}p{(\linewidth - 6\tabcolsep) * \real{0.2000}}
  >{\raggedright\arraybackslash}p{(\linewidth - 6\tabcolsep) * \real{0.2000}}@{}}
\caption{\textbf{Table 8.} Processing durations for 500 evaluation
queries, including the three parsing-fallback cases. Each row summarizes
independently measured per-query intervals. Batch scheduling wait and
browser/network overhead outside the experimental pipeline are
excluded.}\tabularnewline
\toprule\noalign{}
\begin{minipage}[b]{\linewidth}\raggedright
Timed scope
\end{minipage} & \begin{minipage}[b]{\linewidth}\raggedright
Median (s)
\end{minipage} & \begin{minipage}[b]{\linewidth}\raggedright
90th percentile (s)
\end{minipage} & \begin{minipage}[b]{\linewidth}\raggedright
95th percentile (s)
\end{minipage} \\
\midrule\noalign{}
\endfirsthead
\toprule\noalign{}
\begin{minipage}[b]{\linewidth}\raggedright
Timed scope
\end{minipage} & \begin{minipage}[b]{\linewidth}\raggedright
Median (s)
\end{minipage} & \begin{minipage}[b]{\linewidth}\raggedright
90th percentile (s)
\end{minipage} & \begin{minipage}[b]{\linewidth}\raggedright
95th percentile (s)
\end{minipage} \\
\midrule\noalign{}
\endhead
\bottomrule\noalign{}
\endlastfoot
Parallel preparation and recall & 2.69 & 3.30 & 3.56 \\
Gemini reranking & 10.68 & 14.32 & 16.39 \\
Complete query processing & 13.37 & 17.14 & 19.36 \\
\end{longtable}

\begin{figure}
\centering
\includegraphics[width=0.88\linewidth,height=\textheight,keepaspectratio,alt={Empirical cumulative distributions of processing time for 500 evaluation queries. No observations are filtered or smoothed. Preparation includes parallel extraction and rewriting plus local recall; the full interval also includes reranking and local recording.}]{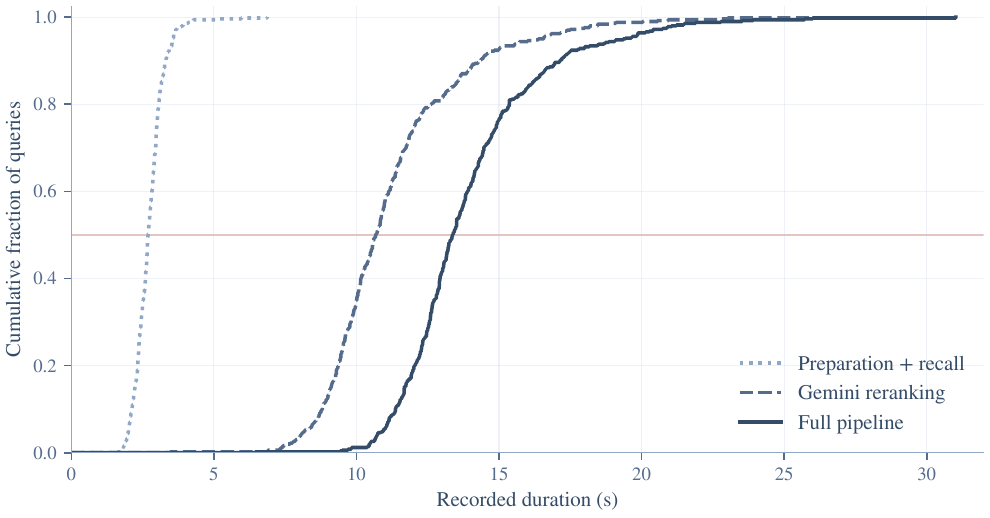}
\caption{Empirical cumulative distributions of processing time for 500
evaluation queries. No observations are filtered or smoothed.
Preparation includes parallel extraction and rewriting plus local
recall; the full interval also includes reranking and local
recording.}\label{fig:latency}
\end{figure}

The 1,000 DeepSeek calls report 863,291 total tokens, and the 500 Gemini
calls report 2,965,848. These are API-reported input-plus-output counts;
reasoning tokens already included in output are not counted again.
Appendix B separates inputs and outputs. Token usage and elapsed time
are tied to the recorded endpoints and concurrency, so no monetary or
browser-latency estimate is inferred from them.

Extraction responses for queries 48, 55, and 489 could not be parsed.
Each query used the original-query fallback and a successful rewrite;
all three targets still ranked first. All 500 rerankings supplied
complete scores. A separate post hoc audit found the additional
static-image caution in 14/500 actual prompts. After whitespace
normalization, a literal text check found a multi-character target
meaning in the truncated target-candidate text for 111 of the 488
queries whose target entered the window. Neither finding proves how
Gemini used these instructions or cues. All cases remain in the results.

\section{7 Practical Value and Trial
Use}\label{practical-value-and-trial-use}

SignTrace is designed for a concrete need in sign-language learning:
identifying a sign whose movement is remembered but whose meaning is
unknown. A learner can describe handshape, contact, and movement in
ordinary words, then inspect the returned dictionary entries,
illustrations, and meanings. This gives partial observations a usable
search path and allows the learner to check whether a candidate fits the
remembered action. Conventional word-to-sign lookup remains available
for revisiting known vocabulary.

The system has been deployed as a working service for user trials and
has received positive informal feedback. Learners can use the movement
details they remember to find signs documented in the dictionary. The
feedback is an author-reported early observation; this evaluation
includes no recruitment protocol, participant-level outcomes, or
controlled usability measurements. Section 6 reports retrieval accuracy
on the dictionary-derived benchmark. The deployed service makes the same
lookup workflow available to users.

\section{8 Discussion}\label{discussion}

\subsection{8.1 Architectural
contributions}\label{architectural-contributions}

SignTrace connects an observer's everyday vocabulary with the action
details needed for dictionary search. LLMs enrich entries with visual
information, express incomplete observations through complementary query
representations, and compare the retrieved candidates. The output
remains a set of dictionary entries with illustrations and meanings,
giving learners material they can inspect when identifying an unfamiliar
sign.

The component analyses support this architecture. Across the evaluation
dataset, raw-query BM25 retrieves 249 targets first using source action
descriptions and 333 after adding visual re-descriptions and structured
summaries. Gemini reranking raises the complete system's first-result
hits from 359 to 470. Enrichment improves access to relevant entries,
while reranking improves the order of candidates that reach its window.
A correct first result provides an immediately inspectable entry, and
the retained shortlist supports comparison when the first candidate is
unsuitable. The channel controls further distinguish initial Hit@1 from
the coverage needed to make a target available to the reranker.

\subsection{8.2 Limitations}\label{limitations}

The benchmark is dictionary-derived, so its wording and action details
may resemble the indexed descriptions. This dependence can favor methods
that use related annotations. Prior retrieval-based selection of 200
queries introduces an additional selection bias. Consequently, the
reported results characterize performance on this benchmark and do not
establish generalization to descriptions independently produced by
users.

Each query has one designated entry. The benchmark neither assesses all
acceptable alternatives nor tests signs outside the dictionary.
Model-generated annotations may contain errors, and historical
annotation and human-review records are incomplete. No independent
accuracy estimate for those annotations is available. The test also
contains no controlled comparisons across signers, dialects, video
conditions, or levels of sign-language experience.

The current endpoint responses are not immutable model snapshots. A
single execution per query measures neither repeated-call stability nor
future behavior after a backend update. All first-stage controls are
exploratory and share the same recorded model transformations. No
comparison across the 500 queries tested a shared reranker, dense
retrieval, alternative paraphrase styles, or an enlarged reranking
window. Candidate-gloss removal and correction of the inherited
image-oriented prompt have not been evaluated; their effects on the
reported rankings remain unmeasured. Action-match scores are ranking
signals and have not been calibrated as confidence probabilities.

\subsection{8.3 Future work}\label{future-work}

Further evaluation should use descriptions independently written by
observers who have not seen the indexed text. Deaf signers and
sign-language educators should participate in study design and
assessment \citep{bestpractice2025}. It should measure lookup success,
time to identify an entry, acceptable alternatives, and errors arising
from incomplete descriptions. Candidate-window expansion should be
tested on the currently missed targets and on a broader fixed sample,
with accuracy, latency, and cost reported together.

The architecture can be adapted to another sign language by supplying
its own dictionary, action descriptions, tokenization, feature
vocabularies, and prompts. Sign languages differ in form, usage, and
regional variation; the written language used for a query is a separate
choice. The current experiments cover the Chinese dictionary evaluated
here only. Performance on another resource requires a new evaluation.

\section{9 Conclusion}\label{conclusion}

SignTrace provides a practical solution to the longstanding difficulty
of identifying a sign from its movement when its meaning is unknown. The
architecture integrates LLM-based dictionary enrichment, action
extraction, query rewriting, and reranking with seven-channel retrieval.
Evaluation on 500 dictionary-derived queries over 6,699 entries yields
94.0\% Hit@1, 97.4\% Hit@9, and a mean reciprocal rank of 0.9540.

The component analyses show the contribution of enriched descriptions
and candidate reranking, while failure analysis identifies limits in
both candidate selection and action comparison. The service has already
been released for user trials and has received positive informal
feedback. The deployed system makes movement-based dictionary lookup
available in practice, and the experiments identify the components that
improve retrieval. Independent user studies remain necessary to measure
lookup success and learning benefits under everyday observation
conditions.

\section{Declarations}\label{declarations}

\textbf{Corresponding author.} Zengji Tu, Peking University;
tuzengji@stu.pku.edu.cn.

\textbf{Author contributions.} Zengji Tu, Xingye Zhu, Ningjing Wang, and
Tingyi Huang contributed to technical development. These four authors
and Yangjunfeng Zhu contributed to system testing and iterative
evaluation. All authors contributed to evaluation-data preparation. Dai
Wan provided supervision and technical guidance.

\textbf{Funding and competing interests.} This research received no
external funding; all expenses were covered by the authors. The authors
declare no financial or non-financial competing interests.

\textbf{Human-participant data.} The quantitative evaluation uses
constructed descriptions and dictionary entries. It reports no
human-participant experiment or participant-level trial data. The
informal trial feedback is described separately in Section 7.

\textbf{Data and code.} Supplementary material includes the evaluation
queries and target identifiers, rankings and processing times,
statistical and visualization scripts, and the evaluated prompt
templates and scoring rules. These materials support reproduction of the
reported analyses and inspection of the retrieval method. The source
dictionary, original illustrations, full annotations, and private API
logs are not redistributed. End-to-end replication requires access to
the dictionary corpus and model services.

\section{Appendix A Statistical
Methods}\label{appendix-a-statistical-methods}

For two methods on the same \(n\) queries, let \(w\) count queries
correct only for the first method and \(l\) those correct only for the
second. The observed accuracy difference is \((w-l)/n\). The exact
two-sided comparison conditions on \(w+l\) discordant queries and tests
a success probability of \(1/2\). Correctness ties remain in the
accuracy denominator.

The difference intervals in Table 5 use a conservative construction
suitable for sparse discordant pairs. We obtain separate 97.5\%
two-sided Clopper--Pearson intervals \([L_w,U_w]\) and \([L_l,U_l]\) for
the win and loss probabilities among all \(n\) queries. The interval
\([L_w-U_l,U_w-L_l]\) then has at least 95\% simultaneous coverage by
the union bound under the query-level binomial sampling model. This
retains uncertainty for an unobserved loss when only a few wins occur.
Holm correction controls the four displayed exact tests as one
comparison family, not the broader development history.

These intervals and tests characterize uncertainty conditional on the
evaluation dataset. They do not account for dependence induced by shared
dictionary content or establish representative sampling of the user
population.

\FloatBarrier
\Needspace{24\baselineskip}

\section{Appendix B Experimental Configuration and Model
Usage}\label{appendix-b-experimental-configuration-and-model-usage}

\FloatBarrier

\begin{longtable}[]{@{}
  >{\raggedright\arraybackslash}p{(\linewidth - 2\tabcolsep) * \real{0.2800}}
  >{\raggedright\arraybackslash}p{(\linewidth - 2\tabcolsep) * \real{0.7200}}@{}}
\caption{\textbf{Table B1.} Experimental configuration for the 500-query
evaluation. Unspecified model sampling parameters retain the endpoint
defaults. The requested and returned strings are recorded identifiers,
not independent verification of underlying model
weights.}\tabularnewline
\toprule\noalign{}
\begin{minipage}[b]{\linewidth}\raggedright
Component
\end{minipage} & \begin{minipage}[b]{\linewidth}\raggedright
Recorded setting
\end{minipage} \\
\midrule\noalign{}
\endfirsthead
\toprule\noalign{}
\begin{minipage}[b]{\linewidth}\raggedright
Component
\end{minipage} & \begin{minipage}[b]{\linewidth}\raggedright
Recorded setting
\end{minipage} \\
\midrule\noalign{}
\endhead
\bottomrule\noalign{}
\endlastfoot
Execution date & 2026-09-13; 500 queries evaluated \\
Dictionary & 6,699 entries; 8,687 listed meanings \\
Text retrieval & BM25; Chinese characters and adjacent character pairs;
\(k_1=1.5\), \(b=0.75\) \\
Fusion & T, Td, Tmv, Traw, Tq2d, S, G; \(c=300\), \(m=3\), \(w=1.5\) \\
Channel voting limits & Positive scores for T/Td/S; first 20 for
Tmv/Traw/Tq2d; first 10 for G \\
Candidate pool and window & 60 retrieved entries; first 20 reranked \\
Candidate action text & At most 200 characters; explicit sense list
omitted \\
Final order & Gemini scores 0--100; score ties preserve recall order;
tail unchanged \\
Query processing & DeepSeek extraction and rewriting in parallel \\
DeepSeek requested / returned & \texttt{deepseek-v4-flash} /
\texttt{deepseek-flash} \\
Gemini requested / returned & \texttt{gemini-3.8-flash} /
\texttt{gemini-3.8-flash} \\
Output token caps & Extraction 4,000; rewriting 600; reranking 16,000 \\
Concurrency and schedule & Six active queries; shuffled with seed
20260913 \\
Cache and failure handling & Application model-output caches disabled;
parsing fallbacks retained; no alternate reranker \\
\end{longtable}

Table B1 reports the requested and returned API model identifiers. These
identifiers may refer to mutable service endpoints rather than fixed
model weights. Disabling application-level caches does not exclude
provider-side prompt caching. The supplementary configuration records
document the model settings used in the evaluation.

\FloatBarrier
\Needspace{12\baselineskip}

\begin{longtable}[]{@{}
  >{\raggedright\arraybackslash}p{(\linewidth - 8\tabcolsep) * \real{0.3200}}
  >{\raggedright\arraybackslash}p{(\linewidth - 8\tabcolsep) * \real{0.0800}}
  >{\raggedright\arraybackslash}p{(\linewidth - 8\tabcolsep) * \real{0.2000}}
  >{\raggedright\arraybackslash}p{(\linewidth - 8\tabcolsep) * \real{0.2000}}
  >{\raggedright\arraybackslash}p{(\linewidth - 8\tabcolsep) * \real{0.2000}}@{}}
\caption{\textbf{Table B2.} API-reported token usage for all 1,500
calls. Output counts already include any reasoning tokens counted within
completion usage. No monetary conversion is applied.}\tabularnewline
\toprule\noalign{}
\begin{minipage}[b]{\linewidth}\raggedright
Model role
\end{minipage} & \begin{minipage}[b]{\linewidth}\raggedright
Calls
\end{minipage} & \begin{minipage}[b]{\linewidth}\raggedright
Input tokens
\end{minipage} & \begin{minipage}[b]{\linewidth}\raggedright
Output tokens
\end{minipage} & \begin{minipage}[b]{\linewidth}\raggedright
Total tokens
\end{minipage} \\
\midrule\noalign{}
\endfirsthead
\toprule\noalign{}
\begin{minipage}[b]{\linewidth}\raggedright
Model role
\end{minipage} & \begin{minipage}[b]{\linewidth}\raggedright
Calls
\end{minipage} & \begin{minipage}[b]{\linewidth}\raggedright
Input tokens
\end{minipage} & \begin{minipage}[b]{\linewidth}\raggedright
Output tokens
\end{minipage} & \begin{minipage}[b]{\linewidth}\raggedright
Total tokens
\end{minipage} \\
\midrule\noalign{}
\endhead
\bottomrule\noalign{}
\endlastfoot
DeepSeek: two query transformations & 1000 & 638,068 & 225,223 &
863,291 \\
Gemini: candidate reranking & 500 & 2,242,676 & 723,172 & 2,965,848 \\
\end{longtable}

\FloatBarrier
\Needspace{12\baselineskip}

\section{Appendix C Supplementary Retrieval
Results}\label{appendix-c-supplementary-retrieval-results}

\FloatBarrier

\begin{longtable}[]{@{}
  >{\raggedright\arraybackslash}p{(\linewidth - 12\tabcolsep) * \real{0.1300}}
  >{\raggedright\arraybackslash}p{(\linewidth - 12\tabcolsep) * \real{0.1450}}
  >{\raggedright\arraybackslash}p{(\linewidth - 12\tabcolsep) * \real{0.1450}}
  >{\raggedright\arraybackslash}p{(\linewidth - 12\tabcolsep) * \real{0.1450}}
  >{\raggedright\arraybackslash}p{(\linewidth - 12\tabcolsep) * \real{0.1450}}
  >{\raggedright\arraybackslash}p{(\linewidth - 12\tabcolsep) * \real{0.1450}}
  >{\raggedright\arraybackslash}p{(\linewidth - 12\tabcolsep) * \real{0.1450}}@{}}
\caption{\textbf{Table C1.} Recall and final target-hit counts and
percentages at all recorded cutoffs. Every cell uses the complete
500-query denominator.}\tabularnewline
\toprule\noalign{}
\begin{minipage}[b]{\linewidth}\raggedright
Stage
\end{minipage} & \begin{minipage}[b]{\linewidth}\raggedright
Hit@1
\end{minipage} & \begin{minipage}[b]{\linewidth}\raggedright
Hit@3
\end{minipage} & \begin{minipage}[b]{\linewidth}\raggedright
Hit@5
\end{minipage} & \begin{minipage}[b]{\linewidth}\raggedright
Hit@9
\end{minipage} & \begin{minipage}[b]{\linewidth}\raggedright
Hit@20
\end{minipage} & \begin{minipage}[b]{\linewidth}\raggedright
Pool@60
\end{minipage} \\
\midrule\noalign{}
\endfirsthead
\toprule\noalign{}
\begin{minipage}[b]{\linewidth}\raggedright
Stage
\end{minipage} & \begin{minipage}[b]{\linewidth}\raggedright
Hit@1
\end{minipage} & \begin{minipage}[b]{\linewidth}\raggedright
Hit@3
\end{minipage} & \begin{minipage}[b]{\linewidth}\raggedright
Hit@5
\end{minipage} & \begin{minipage}[b]{\linewidth}\raggedright
Hit@9
\end{minipage} & \begin{minipage}[b]{\linewidth}\raggedright
Hit@20
\end{minipage} & \begin{minipage}[b]{\linewidth}\raggedright
Pool@60
\end{minipage} \\
\midrule\noalign{}
\endhead
\bottomrule\noalign{}
\endlastfoot
Recall & 359 (71.8) & 427 (85.4) & 453 (90.6) & 475 (95.0) & 488 (97.6)
& 497 (99.4) \\
Final & 470 (94.0) & 482 (96.4) & 484 (96.8) & 487 (97.4) & 488 (97.6) &
497 (99.4) \\
\end{longtable}

\FloatBarrier
\Needspace{23\baselineskip}

\begin{longtable}[]{@{}
  >{\raggedright\arraybackslash}p{(\linewidth - 6\tabcolsep) * \real{0.0750}}
  >{\raggedright\arraybackslash}p{(\linewidth - 6\tabcolsep) * \real{0.5250}}
  >{\raggedright\arraybackslash}p{(\linewidth - 6\tabcolsep) * \real{0.2000}}
  >{\raggedright\arraybackslash}p{(\linewidth - 6\tabcolsep) * \real{0.2000}}@{}}
\caption{\textbf{Table C2.} All 13 queries whose final target rank
exceeds nine. Absent means the target does not occur in the 60-entry
candidate pool. Targets beyond the 20-entry window retain their recall
rank.}\tabularnewline
\toprule\noalign{}
\begin{minipage}[b]{\linewidth}\raggedright
ID
\end{minipage} & \begin{minipage}[b]{\linewidth}\raggedright
Target meaning
\end{minipage} & \begin{minipage}[b]{\linewidth}\raggedright
Recall rank
\end{minipage} & \begin{minipage}[b]{\linewidth}\raggedright
Final rank
\end{minipage} \\
\midrule\noalign{}
\endfirsthead
\toprule\noalign{}
\begin{minipage}[b]{\linewidth}\raggedright
ID
\end{minipage} & \begin{minipage}[b]{\linewidth}\raggedright
Target meaning
\end{minipage} & \begin{minipage}[b]{\linewidth}\raggedright
Recall rank
\end{minipage} & \begin{minipage}[b]{\linewidth}\raggedright
Final rank
\end{minipage} \\
\midrule\noalign{}
\endhead
\bottomrule\noalign{}
\endlastfoot
14 & Cauliflower & 27 & 27 \\
17 & Dismantle & 49 & 49 \\
18 & Exceed a production target & 50 & 50 \\
38 & Regular; periodic & 56 & 56 \\
58 & Bright; light & Absent & Absent \\
114 & Shallow & 33 & 33 \\
215 & Transport & 28 & 28 \\
230 & Grain Full (solar term) & 29 & 29 \\
248 & Branch (organization) & Absent & Absent \\
258 & At any time & 36 & 36 \\
343 & Numeral; classifier & 16 & 13 \\
351 & Sugar beet & 42 & 42 \\
395 & Bullet & Absent & Absent \\
\end{longtable}

\FloatBarrier
\Needspace{10\baselineskip}

\begin{longtable}[]{@{}
  >{\raggedright\arraybackslash}p{(\linewidth - 8\tabcolsep) * \real{0.1500}}
  >{\raggedright\arraybackslash}p{(\linewidth - 8\tabcolsep) * \real{0.0700}}
  >{\raggedright\arraybackslash}p{(\linewidth - 8\tabcolsep) * \real{0.2000}}
  >{\raggedright\arraybackslash}p{(\linewidth - 8\tabcolsep) * \real{0.3800}}
  >{\raggedright\arraybackslash}p{(\linewidth - 8\tabcolsep) * \real{0.2000}}@{}}
\caption{\textbf{Table C3.} Descriptive length strata partitioning all
500 evaluation queries. Length counts Unicode characters including
punctuation. Hit columns give counts and percentages; intervals refer to
Hit@1. These post hoc strata compare different queries.}\tabularnewline
\toprule\noalign{}
\begin{minipage}[b]{\linewidth}\raggedright
Characters
\end{minipage} & \begin{minipage}[b]{\linewidth}\raggedright
n
\end{minipage} & \begin{minipage}[b]{\linewidth}\raggedright
Hit@1
\end{minipage} & \begin{minipage}[b]{\linewidth}\raggedright
95\% Wilson CI (\%)
\end{minipage} & \begin{minipage}[b]{\linewidth}\raggedright
Hit@9
\end{minipage} \\
\midrule\noalign{}
\endfirsthead
\toprule\noalign{}
\begin{minipage}[b]{\linewidth}\raggedright
Characters
\end{minipage} & \begin{minipage}[b]{\linewidth}\raggedright
n
\end{minipage} & \begin{minipage}[b]{\linewidth}\raggedright
Hit@1
\end{minipage} & \begin{minipage}[b]{\linewidth}\raggedright
95\% Wilson CI (\%)
\end{minipage} & \begin{minipage}[b]{\linewidth}\raggedright
Hit@9
\end{minipage} \\
\midrule\noalign{}
\endhead
\bottomrule\noalign{}
\endlastfoot
13--24 & 45 & 40 (88.9) & 76.5--95.2 & 45 (100.0) \\
25--40 & 307 & 286 (93.2) & 89.8--95.5 & 297 (96.7) \\
41--64 & 148 & 144 (97.3) & 93.3--98.9 & 145 (98.0) \\
\end{longtable}
\clearpage
\bibliographystyle{unsrtnat}
\bibliography{references}
\end{document}